\documentclass[letterpaper]{article} 
\usepackage[preprint]{aaai2027}  
\usepackage[hyphens]{url}  
\usepackage{graphicx} 
\usepackage{natbib}  
\usepackage{caption} 
\usepackage{algorithm}
\usepackage{algpseudocode}
\usepackage{booktabs}
\usepackage{amsmath}
\usepackage{amssymb}
\usepackage{mathtools}
\usepackage{ifpdf}

\usepackage{hyperref} 

\DeclareCaptionStyle{ruled}{labelfont=normalfont,labelsep=colon,strut=off} 
\newcommand{\lossScalingFigure}{%
  \begin{figure*}[!t]
    \centering
    \includegraphics[width=0.98\textwidth]{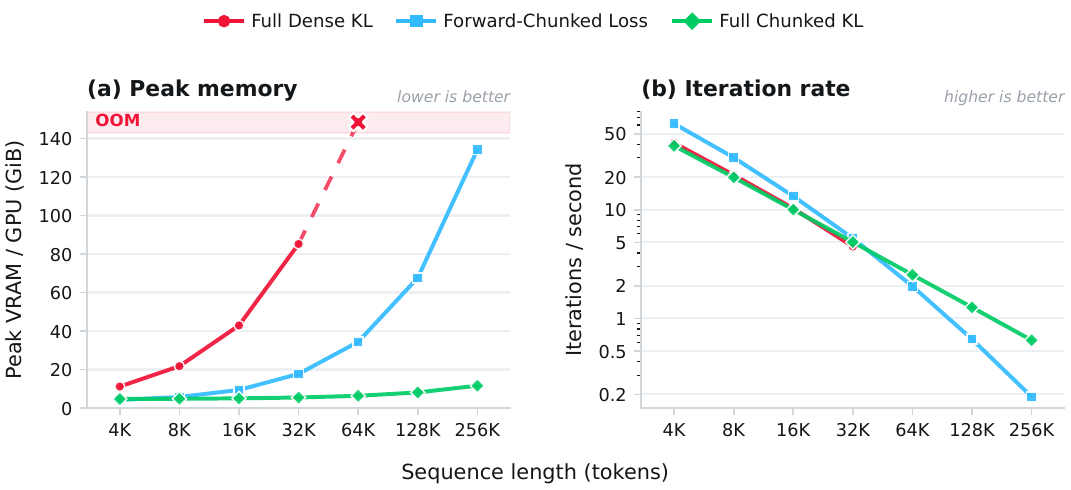}
    \caption{Controlled loss-only benchmark \textbf{Left:}
    peak memory per GPU.
    \textbf{Right:} forward-backward iteration rate on a logarithmic scale.
    Dense KL fails from 64K onward; forward-chunked is fastest through 32K,
    while fully chunked is faster at longer contexts and uses substantially
    less memory. Batch size $1$, hidden size $4{,}096$, vocabulary $131{,}072$,
    top-$K{=}100$, tensor parallelism $2$, and chunk size $4{,}096$.}
    \label{fig:loss-only-scaling}
  \end{figure*}%
}

\title{Efficient Knowledge Distillation for LLMs:\\ Offline Top-K Logits and a Fused Chunked KL Loss}
\author{
Bakbergen Ryskulov\textsuperscript{1},
Iker García-Ferrero\textsuperscript{2},
David Montero,
David Jansen,
Ali Hashemi,
Jezabel R. Garcia,
Antonio Tiene,
Román Orús
}
\affiliations{
    Multiverse Computing \\
   \url{https://multiversecomputing.com} 
  
}

\begin{document}

\maketitle

\begingroup
\renewcommand{\thefootnote}{\arabic{footnote}}
\footnotetext[1]{\fontsize{10.5pt}{9pt}\selectfont
\url{bakbergen.ryskulov@multiversecomputing.com}}
\footnotetext[2]{\fontsize{10.5pt}{9pt}\selectfont
\url{iker.garcia@multiversecomputing.com}}
\endgroup

\begin{abstract}
Small language models are often the only option for deployment under tight
latency, cost, and on-premises constraints, but they are rarely trained from
scratch: a compressed model is usually recovered through knowledge
distillation (KD). This recovery step largely decides the final quality, yet
it is expensive. We present a practitioner's study of how to make distillation
\emph{training} efficient, organised around two systems contributions. First,
we show that \emph{offline} KD (caching the teacher's top-$K$ logits once and
training the student against the cache) matches \emph{online} distillation at
near-identical training loss while removing the teacher from memory, running
about 29\% faster per iteration, and reaching up to 41\% higher throughput on a
single H200 GPU. Second, we introduce a \emph{fused, chunked KL loss} that
never materialises the full vocabulary-sized logit tensor, making peak memory
linear in the sequence length. This removes the memory spike that otherwise
caps context length and lets us train at four times the context (32{,}768
tokens) on a single GPU. A separate output-head-only toy benchmark isolates the
loss kernel and confirms its memory and iteration-rate scaling from 4K to 256K
tokens. Together these make large-scale healing and hundreds of ablations
affordable. We also report supporting ablations on loss design and sequence
packing. We release our chunked-loss implementation:
\url{https://github.com/CompactifAI/Full-Chunked-KL-Loss}.
\end{abstract}

\section{Introduction}

\begin{figure}[t]
  \centering
  \includegraphics[width=1.0\linewidth]{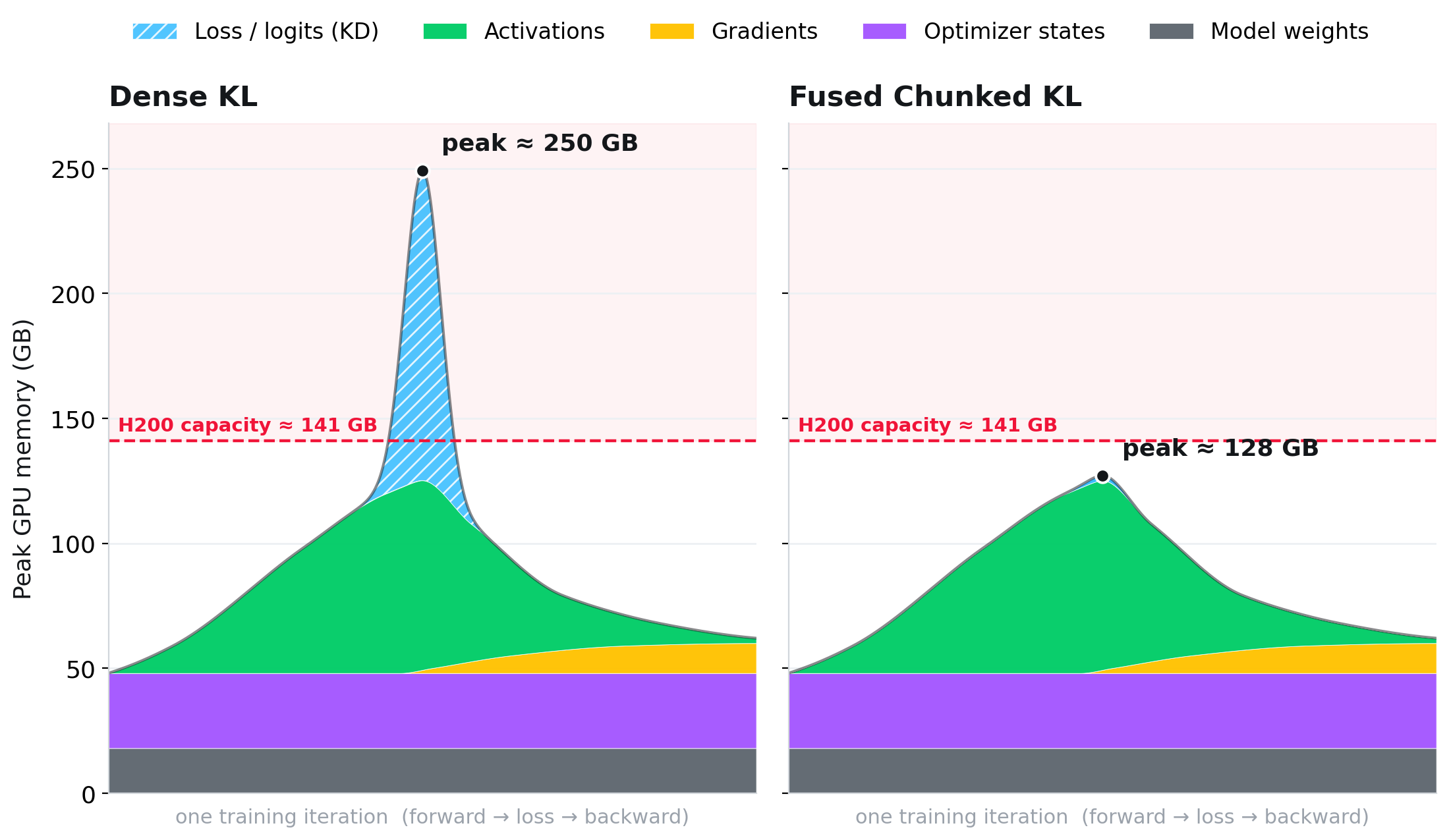}
  \caption{The fused chunked KL loss unlocks long-context healing. GPU memory
    over one training iteration at a 32K context length, broken down by
    component. The dense KL loss materialises a vocabulary-sized logit/teacher
    tensor (hatched, \emph{extrapolated}) that spikes peak memory near
    $250$~GB, exceeding a single H200's $141$~GB capacity, whereas the fused
    chunked loss never forms that tensor and peaks at $128$~GB. Only the
    loss/logits component differs; the remaining components are an estimated
    split of the measured total.}
  \label{fig:teaser}
\end{figure}

\label{sec:intro}
Large language models are increasingly deployed under hard production
constraints: tight latency budgets, per-token cost ceilings, and on-premises
serving where the largest models do not fit. The common response is to deploy
a compact model derived from a capable teacher, whose quality is then
\emph{recovered} through knowledge distillation, training the student to match
the teacher's behaviour. This recovery step decides most of the practical cost
and the final quality, yet it is under-documented relative to its impact.

We report guidance from an extensive distillation campaign on a compact
($\sim$3.2B) student derived from Llama~3.1~8B~Instruct
\citep{dubey2024llama3}. The method used to obtain the compact
initialisation is independent of this work \citep{muralidharan2024minitron};
the \emph{distillation recipe} applies to any compact model initialised from a
larger teacher. This is a practically driven contribution in the strict sense:
no new algorithm, but deployment-driven choices measured at scale and reported
with their trade-offs, including where they fail.

The paper makes two efficiency contributions, which we deliberately keep
separate because they address different bottlenecks.

\paragraph{Offline distillation.} Computing the teacher on the fly
(\emph{online} KD) keeps both models resident and recomputes the teacher
forward pass at every step. We instead compute the teacher logits once, cache
the top-$K$ per token, and train the student against the cache. This matches
online quality at near-identical loss while removing the teacher from memory
and the teacher forward pass from the loop, which lowers cost and, crucially
for a research campaign, lets us run tens of ablations against the same cached
targets (\S\ref{sec:exp-online-offline}).

\paragraph{A fused, chunked KL loss.} The binding constraint on long-context
healing is not the transformer body but the logit tensor produced by the
language-model head and its loss, whose vocabulary-sized footprint spikes peak
memory and caps the trainable sequence length
\citep{wijmans2025cut,hsu2024liger}. Memory-efficient cross-entropy losses that
fuse the output projection and chunk the sequence are by now established
\citep{wijmans2025cut,hsu2024liger}; we bring the technique to a
\emph{knowledge-distillation} objective, where the target is a sparse top-$K$
teacher distribution with partially retained mass rather than a one-hot label,
giving a forward KL with a different closed-form gradient
(\S\ref{sec:fullchunked}). The result makes peak memory linear in the sequence
length (Figure~\ref{fig:teaser}), removing the spike that caps context and
unlocking the long-context healing a compact student needs to serve long inputs
\citep{liu2024lost,gao2025train}. This implementation is not available in
current libraries and we will release it.

\noindent\textbf{Contributions.}
\begin{itemize}
  \item An \emph{offline} top-$K$ logit-distillation pipeline that matches
        online quality while cutting memory and raising throughput, which is
        what makes large-scale ablation studies practical
        (\S\ref{sec:exp-online-offline}).
  \item A \emph{fused, chunked KL loss} that extends memory-efficient
        cross-entropy kernels \citep{wijmans2025cut,hsu2024liger} to a sparse
        top-$K$ teacher, making peak memory linear in the sequence length and
        unlocking long-context healing on a single GPU, with the loss-kernel
        scaling isolated in a controlled output-head benchmark
        (\S\ref{sec:fullchunked}, \S\ref{sec:exp-chunked},
        \S\ref{sec:exp-loss-only}).
  \item Supporting ablations enabled by the efficient setup, (loss design and
        sequence packing) that round out a reproducible recipe
        (\S\ref{sec:exp-additional}).
\end{itemize}

\section{Related Work}
\label{sec:related}

\paragraph{Knowledge distillation.}
Distilling a large teacher into a smaller student is a long-standing route
to compression, from the original logit-matching formulation
\citep{hinton2015distilling} to task-agnostic distillation of pretrained
transformers \citep{sanh2019distilbert,wang2020minilm}. At the scale of
instruction-tuned language models, distillation is most often used as the
\emph{recovery} step after structured pruning: Minitron prunes a large model
and heals it with distillation \citep{muralidharan2024minitron}, and the Gemma
reports distil from top-$K$ teacher logits, with $K$ of order a few hundred
\citep{gemma2024}, which motivates our study of how many cached logits are
actually needed. The closest work in framing is the industry-track comparison
of task-agnostic distillation methods of \citet{udagawa2023comparative}; we
extend that line to instruction-tuned LLM scale, an offline and
memory-efficient training pipeline, and long-context healing with an explicit
deployment framing. Our aim is a practitioner's recovery recipe rather than a
new distillation algorithm, so we report the choices and their trade-offs,
including where they fail.

\paragraph{Long context and memory-efficient losses.}
Long-context ability is now a first-class capability: models are asked to
reason over long documents, retrieved passages, and agentic histories, and use
that context unevenly enough to need careful evaluation \citep{liu2024lost}, so
we adopt HELMET \citep{yen2025helmet} with its Ruler and retrieval-augmented
generation tasks \citep{hsieh2024ruler} to locate where our student regresses.
The same capability must be acquired in training, where the binding cost is
not the transformer body but the logit tensor materialised by the
language-model head and its loss \citep{wijmans2025cut,gao2025train}. Cut
Cross-Entropy \citep{wijmans2025cut} and the Liger kernels \citep{hsu2024liger}
remove this cost for the \emph{cross-entropy} objective by chunking the
sequence and fusing the output projection so the full logits are never
materialised. Our fused chunked loss (\S\ref{sec:fullchunked}) brings the same
technique to a \emph{knowledge-distillation} objective these libraries do not
support: the target is a sparse top-$K$ teacher with retained mass $M\le 1$, so
the loss is a forward KL divergence and the closed-form gradient is a dense
``$M\cdot\mathrm{softmax}$'' term minus a sparse teacher correction
(Equation~\ref{eq:grad}), rather than a single ground-truth subtraction. This
adaptation is the part absent from current libraries, and we will release it.

\section{Types of Knowledge Distillation}
\label{sec:distillation}

We distil a large teacher into a smaller student by matching their output
distributions with a forward Kullback--Leibler (KL) objective. For a single
token, let $z\in\mathbb{R}^{V}$ be the student logits over a vocabulary of size
$V$ and $q_v=\exp(z_v)/Z$ with $Z=\sum_{u}\exp(z_u)$ the student probability of
token $v$, so that $\log q_v = z_v - \log Z$. With $p\in\mathbb{R}^{V}$ the
teacher distribution, the per-token loss is
\begin{equation}
  \mathcal{L}_{\mathrm{KL}}(p,z)
  = \sum_{v=1}^{V} p_v\,(\log p_v - \log q_v),
  \label{eq:kl}
\end{equation}
averaged over all next-token-shifted, loss-masked positions. We consider two
regimes for obtaining $p$: \emph{online}, where the teacher is resident in
memory, and \emph{offline}, where only the teacher's top-$K$ probabilities are
precomputed and cached.

\subsection{Online Distillation}
\label{sec:online}
In the online setting both teacher and student are loaded simultaneously. A
teacher forward pass produces the dense distribution $p\in\mathbb{R}^{V}$ for
every position, and \eqref{eq:kl} is evaluated directly against the student's
dense log-softmax. This regime is the most expressive (the full teacher
distribution is available) but the most memory- and compute-intensive: it holds
both models and materialises two dense $\mathbb{R}^{V}$ tensors per position
(teacher probabilities and student log-probabilities), on top of recomputing
the teacher at every step.

\subsection{Offline Distillation with a Top-$K$ Teacher}
\label{sec:offline-kl}
To remove the teacher from memory, we precompute and cache only its $K{=}100$
largest probabilities per position. Let $\mathcal{S}\subset\{1,\dots,V\}$,
$|\mathcal{S}|=K$, denote this support, with $p_v=0$ for $v\notin\mathcal{S}$.
The retained mass $M=\sum_{v\in\mathcal{S}}p_v\le 1$ may be strictly below one
because of truncation; we do \emph{not} renormalise, and the formulation below
accounts for the partial mass exactly. Substituting $\log q_v = z_v-\log Z$ into
\eqref{eq:kl} and restricting to the support yields the identity that underlies
all three offline implementations:

\begin{equation}
\mathcal{L}_{\mathrm{KL}}(p,z)
=
\underbrace{\smashoperator{\sum_{v\in\mathcal{S}}} p_v\log p_v}_{H}
-
\underbrace{\smashoperator{\sum_{v\in\mathcal{S}}} p_v z_v}_{C}
+ M\log Z .
\label{eq:decomp}
\end{equation}
The teacher-entropy term $H$, the cross term $C$, and the mass $M$ depend only
on the $K$ support entries, so they require gathering just $K$ student logits
per position. Only the log-normaliser $\log Z$ depends on the entire
vocabulary, but it is a \emph{scalar per position}, a reduction, not an
$\mathbb{R}^{V}$ tensor. The three offline methods below are mathematically
equivalent evaluations of \eqref{eq:decomp}; they differ only in how they
handle $\log Z$ and the student logits. We use the untempered objective
($\tau=1$) throughout.

\subsubsection{Full Dense KL Computation}
\label{sec:fullkl}
The simplest approach reconstructs the dense teacher: the cached top-$K$ values
are scattered into a dense tensor $p\in\mathbb{R}^{B\times S\times V}$ and the
objective is evaluated against the student's dense log-softmax, exactly as in
the online case. It therefore materialises two vocabulary-sized tensors (the
reconstructed top-$K$ teacher and the student log-probabilities) on top of the
student logits, so peak memory is $O(SBV)$. It serves as a correctness baseline:
it is the offline computation closest to online distillation.

\subsubsection{Sparse KL and Forward-Chunked Loss}
\label{sec:chunkedforward}
This variant keeps the teacher sparse and evaluates \eqref{eq:decomp} directly,
never forming a dense teacher or a dense log-softmax. The student logits $z$
(produced by the language-model head) are processed in contiguous chunks of
$C_s$ sequence positions: each chunk computes a numerically stable maximum and
exponential sum over the vocabulary and writes the scalar $\log Z$ for its
positions, while the sparse terms $H$, $C$, and $M$ are accumulated by
scatter-add over the $K$ retained entries. Because the chunking removes only the
\emph{auxiliary} dense tensors, the student logits and their gradient are still
held in full, so peak memory retains a vocabulary-sized $O(SBV)$ term and the
method does not on its own enable longer sequences. It is, however, the fastest
variant in our profiling, since it keeps the standard output projection but
removes the dense teacher, dense log-softmax, and dense KL arithmetic.

\subsubsection{Full Chunked KL Computation}
\label{sec:fullchunked}
This is our main contribution. The loss \emph{fuses the output projection into
the loss}, so the full $[S,B,V]$ logit tensor is never materialised, in neither
the forward pass nor as a stored gradient, only a transient chunk of logits
exists at a time. Given hidden states $h\in\mathbb{R}^{S\times B\times d}$ and
the output projection $W\in\mathbb{R}^{V\times d}$, the forward pass processes
the sequence chunk by chunk: it projects $z_{[s_0:s_1]}=h_{[s_0:s_1]}W^{\top}$,
accumulates $\log Z$ and the sparse terms of \eqref{eq:decomp}, and immediately
discards each chunk of logits. It retains for the backward pass only $h$, the
per-position scalars $\log Z$ and $M$ (each $\mathbb{R}^{S\times B}$), and the
sparse teacher entries. The backward pass recomputes the logits chunk by chunk,
rebuilds $q_{[s_0:s_1]}=\exp(z_{[s_0:s_1]}-\log Z_{[s_0:s_1]})$ from the saved
normaliser, and forms the logit gradient in closed form,
\begin{equation}
  \frac{\partial \mathcal{L}_{\mathrm{KL}}}{\partial z_v}
  = M\,q_v - p_v ,
  \label{eq:grad}
\end{equation}
i.e.\ a dense ``$M\cdot\mathrm{softmax}$'' term minus the sparse teacher
correction at the $K$ support positions. Each chunk gradient is projected back
to accumulate $\partial\mathcal{L}/\partial h$ and $\partial\mathcal{L}/\partial
W$, after which the chunk is freed (Algorithm~\ref{alg:fullchunked}). The
vocabulary-sized cost is thus confined to a single chunk and is independent of
sequence length; the only quantity that grows with $S$ is the hidden-state
activation, $O(SBd)$ with $d\ll V$. Peak memory is therefore \emph{linear in the
sequence length}, in contrast to the $O(SBV)$ footprint of the other two
variants, at the cost of one extra output projection per chunk in the
backward pass, a gradient-checkpointing trade-off on the output head. The
formulation is compatible with vocabulary sharding: the normaliser and sparse
top-$K$ terms are reduced across shards, and each rank stores and
differentiates only its local vocabulary slice.

\begin{algorithm}[t]
\caption{Fused chunked KL (output projection fused into the loss)}
\label{alg:fullchunked}
\begin{algorithmic}[1]
\Require hidden states $h\in\mathbb{R}^{S\times B\times d}$;
  projection $W\in\mathbb{R}^{V\times d}$;
  support $(t,b,v,p)$; chunk size $C_s$
\Statex \textbf{Forward} (no gradient tracked):
\For{each chunk $[s_0,s_1)$} \Comment{Pass 1: normaliser}
  \State $z \gets h_{[s_0:s_1]}W^{\top}$;\quad
         $m \gets \max_v z$ \Comment{distributed max if sharded}
  \State $\sigma_{[s_0:s_1]} \gets \sum_v \exp(z - m)$;\quad discard $z$
\EndFor
\State $\log Z \gets m + \log\sigma$ \Comment{reduce $\sigma$ if sharded}
\For{each chunk $[s_0,s_1)$} \Comment{Pass 2: loss}
  \State $z \gets h_{[s_0:s_1]}W^{\top}$
  \State $M \gets \sum p$;\; $H \gets \sum p\log p$;\; $C \gets \sum p\,z_{t,b,v}$
  \State $\mathcal{L}_{[s_0:s_1]} \gets H - C + M\odot\log Z_{[s_0:s_1]}$;\;
         discard $z$
\EndFor
\State save $h,\,W,\,\log Z,\,M,\,(t,b,v,p)$;\quad \Return $\mathcal{L}$
\Statex \textbf{Backward} given $g=\partial\mathcal{L}/\partial\ell$:
\For{each chunk $[s_0,s_1)$}
  \State $z \gets h_{[s_0:s_1]}W^{\top}$ \Comment{recompute logits}
  \State $q \gets \exp(z - \log Z_{[s_0:s_1]})$ \Comment{rebuild softmax}
  \State $G \gets (M\odot g)_{[s_0:s_1]}\cdot q$
  \State $G_{t,b,v} \mathrel{-}= p\cdot g_{t,b}$ \Comment{Eq.~\eqref{eq:grad}}
  \State $\partial h_{[s_0:s_1]} \gets G\,W$;\quad
         $\partial W \mathrel{+}= G^{\top} h_{[s_0:s_1]}$
\EndFor
\State reduce $\partial h$ across shards;\quad
       \Return $\partial h,\ \partial W$
\end{algorithmic}
\end{algorithm}

\section{Experiments}
\label{sec:experiments}

We organise the experiments around the two contributions and a set of
supporting ablations, and describe the relevant setup inline in each subsection
rather than in a separate section. Unless stated otherwise the teacher is
Llama~3.1~8B~Instruct and the student is a compact $\sim$3.2B model;
distillation uses NVIDIA Megatron-Bridge with the Megatron-LM backend and
NVIDIA ModelOpt, and efficiency is measured with the PyTorch memory profiler,
Megatron-Bridge profiling, and NVIDIA Nsight Systems. The complete training
configuration and software stack are listed in Appendix~\ref{app:config}. The
controlled experiment in \S\ref{sec:exp-loss-only} is the sole exception: it
uses a toy output-projection network and synthetic tensors to isolate the loss
kernel, rather than a pretrained or end-to-end language model.

\begin{figure*}[t]
  \centering
  \includegraphics[width=0.98\textwidth]{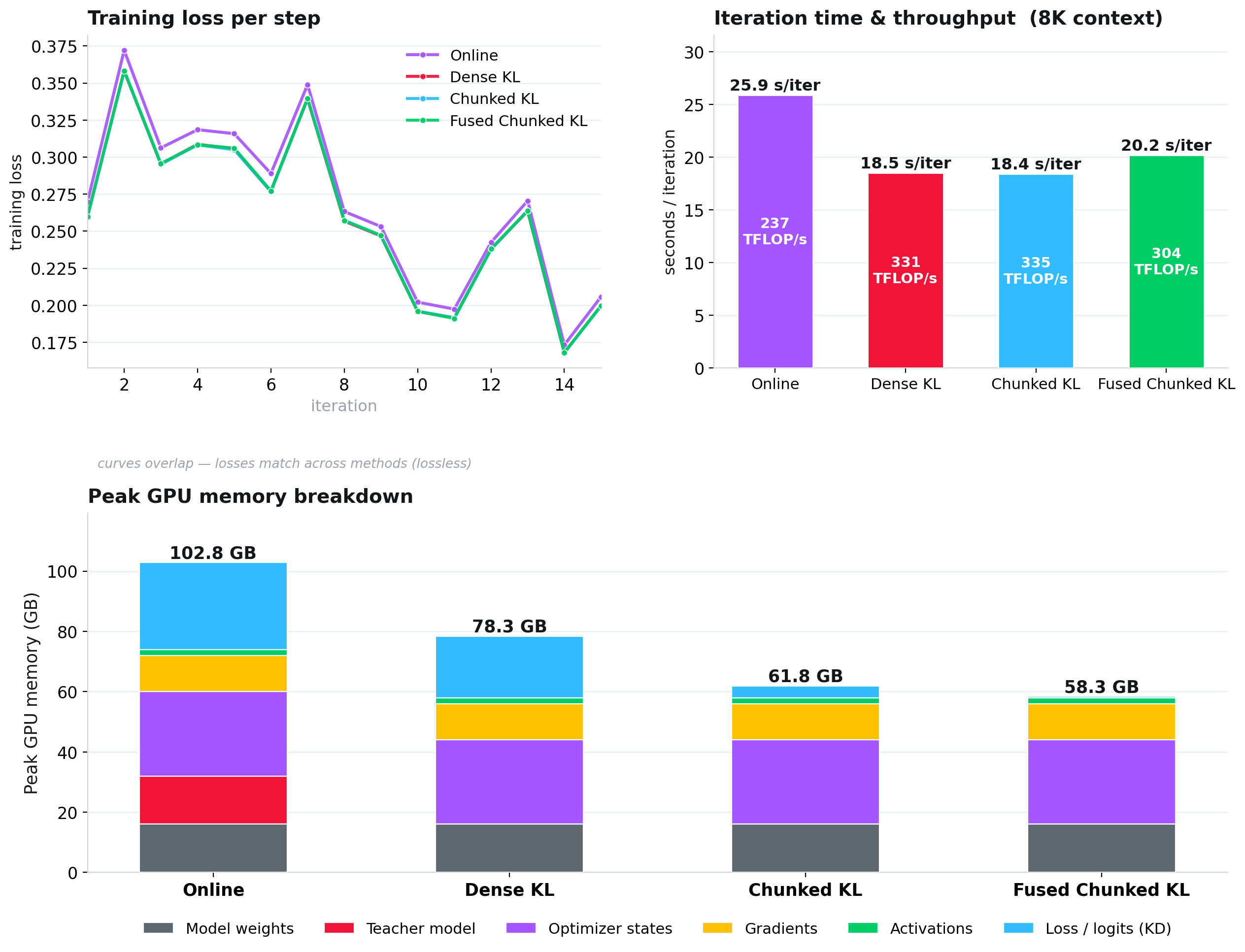}
  \caption{All methods compared at 8K context on a single H200: online
    distillation and the three offline implementations (dense, forward-chunked,
    and fused chunked KL). \textbf{Top-left:} training loss per step is
    near-identical across all methods, including the offline runs that use only
    the top-$100$ cached logits. \textbf{Top-right:} iteration time (bar height)
    and throughput (TFLOP/s, labelled inside each bar). \textbf{Bottom:} peak
    GPU memory by component; offline removes the resident teacher (red), and the
    loss/logits term (blue) shrinks as the vocabulary-sized tensor is chunked
    away, effectively vanishing for the fused loss.}
  \label{fig:all-methods}
\end{figure*}

\subsection{Online vs.\ Offline (Full Dense KL)}
\label{sec:exp-online-offline}
\textbf{Setup.} We profile a single training step on one H200 GPU at a sequence
length of 8{,}192 on the SmolTalk supervised fine-tuning data
\citep{allal2025smollm2}, comparing online
distillation against offline distillation with the full dense KL loss. The
offline run caches the teacher's top-$100$ logits per token; correctness is
assessed by matched training loss, and efficiency by peak memory, throughput,
and seconds per iteration.

\textbf{Results.} The two regimes reach near-identical training-loss curves,
even though the offline run trains against only the top-$100$ cached logits
(Figure~\ref{fig:all-methods}, top-left). Offline distillation is meanwhile
markedly cheaper: it removes the teacher from memory, lowering peak memory from
about $103$ to $78$~GB, runs about $29\%$ faster per iteration
($25.9\rightarrow18.5$~s), and raises throughput from $237$ to $331$~TFLOP/s
($\sim40\%$). The quality match at lower cost is what makes offline
distillation attractive at scale: the teacher need only be run once, after
which the cache can be reused across many ablations, and the savings grow with
teacher size (e.g.\ a 70B teacher need not sit in the training loop).

\ifpdf
  \lossScalingFigure
\fi

\subsection{Adding the Chunked KL Loss}
\label{sec:exp-chunked}
\textbf{Setup.} Using the same single-H200, 8K, SmolTalk profiling setup, we
compare the three offline implementations of the same objective: the full dense
KL baseline (\S\ref{sec:fullkl}), the forward-chunked loss
(\S\ref{sec:chunkedforward}), and the fused chunked loss
(\S\ref{sec:fullchunked}).
\textbf{Results.} All three implementations produce the same training-loss
curve (Figure~\ref{fig:all-methods}, top-left), confirming they evaluate the
same objective. They differ in cost. The fused chunked loss reduces peak memory
furthest, from $78$~GB (dense) to $62$~GB (forward-chunked) to $58$~GB
(fused), by never materialising the vocabulary-sized logit tensor, a reduction
not available in current libraries such as ModelOpt. Because peak memory is now
linear in sequence length rather than $O(SBV)$, the fused loss removes the
memory spike that otherwise caps context: on the same single H200 it trains at
$32{,}768$ tokens, roughly four times the context that fits with the dense loss.
The forward-chunked loss is the fastest per iteration, while the fused loss pays
a small recompute cost in the backward pass; in exchange it is the only variant
that unlocks long context. We stress that under this setting, a single H200 at
8K context, the fused chunked loss is \emph{not} the fastest approach: its extra
output projection in the backward pass makes the forward-chunked loss lead on
iteration time.

\textbf{Scaling to larger models and longer context.} The advantage of the
fused loss grows sharply with model and context size, where the freed memory
turns into a throughput win. In further experiments distilling GPT-OSS-20B at a
context length of $32{,}768$ on $8\times$H200 nodes, the memory it frees lets
the model drop from four nodes (tensor parallel $4$, pipeline parallel $4$,
expert parallel $2$) to a single node (tensor parallel $2$, pipeline parallel
$1$, expert parallel $4$), removing most of the inter-node communication.
Step time then falls from $57.0$ to $12.23$ seconds ($\sim$5$\times$ faster) and
throughput rises from $74.2$ to $345.7$ TFLOP/s per GPU: never instantiating the
$O(S\,B\,V)$ logit tensor both removes the memory bottleneck and pushes GPU
utilisation far higher. The peak-memory reductions reported above
($78\!\rightarrow\!62\!\rightarrow\!58$~GB) are measured at 8K context, and the
gap widens with length. At $32{,}768$ tokens the dense loss peaks at roughly
$250$~GB, beyond a single H200's capacity, so it does not fit, whereas the
fused chunked loss peaks at about $128$~GB.

\subsection{Isolating Loss-Kernel Scaling}
\label{sec:exp-loss-only}
\ifpdf\else
  \lossScalingFigure
\fi

The preceding results measure real LLM training and therefore mix the cost of
the KL implementation with transformer layers, attention, optimiser state,
data movement, and framework overhead. To isolate the mechanism behind the
memory reduction, we additionally run a controlled microbenchmark using a
\emph{toy neural network}. The results are depicted in Figure \ref{fig:loss-only-scaling}. This experiment is deliberately not an LLM
benchmark and its absolute memory and iteration-rate values should not be
compared directly with Figures~\ref{fig:all-methods} and~\ref{fig:teaser}.

\textbf{Setup.} The toy network contains only a vocabulary output projection:
synthetic hidden states $h$ are multiplied by a synthetic weight matrix $W$,
then one of the three KL implementations executes its forward and backward
passes. There are no transformer blocks, attention layers, optimiser step,
data loader, or resident teacher model. Deterministic hidden states, projection
weights, and sparse top-$100$ teacher targets are reused across methods. The
plotted run uses hidden size $4{,}096$, vocabulary size $131{,}072$, batch size
$1$, bfloat16, tensor parallelism $2$, and a $4{,}096$-token chunk for both
chunked variants, over sequence lengths from 4K to 256K. Each configuration is
launched in a fresh distributed subprocess so an out-of-memory failure cannot
contaminate later measurements. Peak allocated CUDA memory and mean
forward-plus-backward iteration time are reduced by the maximum over the two
ranks. Separate deterministic CPU tests verify agreement of the per-token loss
and hidden-state gradients to $10^{-4}$ tolerance. Full details appear in
Appendix~\ref{app:toy-benchmark}.

\textbf{Results.} At 32K tokens, peak memory is $85.2$~GiB for dense KL,
$17.7$~GiB for the forward-chunked loss, and $5.45$~GiB for the fully chunked
loss (Figure~\ref{fig:loss-only-scaling}, left), a $15.6\times$ reduction from
dense to fully chunked. Dense KL then fails at 64K. At 256K, the
forward-chunked loss still holds the full logits and reaches $134.2$~GiB per
GPU, whereas the fully chunked loss uses $11.6$~GiB. The timing panel (Figure~\ref{fig:loss-only-scaling}, right) exposes
the recomputation trade-off: at 32K, forward-chunked leads at $5.46$
iterations/s versus $5.04$ for fully chunked, but the ranking reverses at 64K.
At 256K the fully chunked loss reaches $0.630$ iterations/s versus $0.190$ for
forward-chunked, a $3.3\times$ advantage in this isolated workload.

The fully chunked loss uses $15.6\times$ less
memory than dense KL at 32K and $11.6\times$ less than forward-chunked at 256K.
Although forward-chunked is faster at smaller contexts, fully chunked overtakes
it from 64K onward and is $3.3\times$ faster at 256K. Thus the microbenchmark
validates the intended kernel-level scaling: full-sequence
logits dominate the other implementations, while the fused implementation
bounds vocabulary-sized storage by the chunk.

\subsection{Additional Ablations}
\label{sec:exp-additional}
The efficient offline setup made several smaller studies cheap to run. We
summarise the two that bear directly on the recipe.

\begin{figure}[t]
  \centering
  \includegraphics[width=\linewidth]{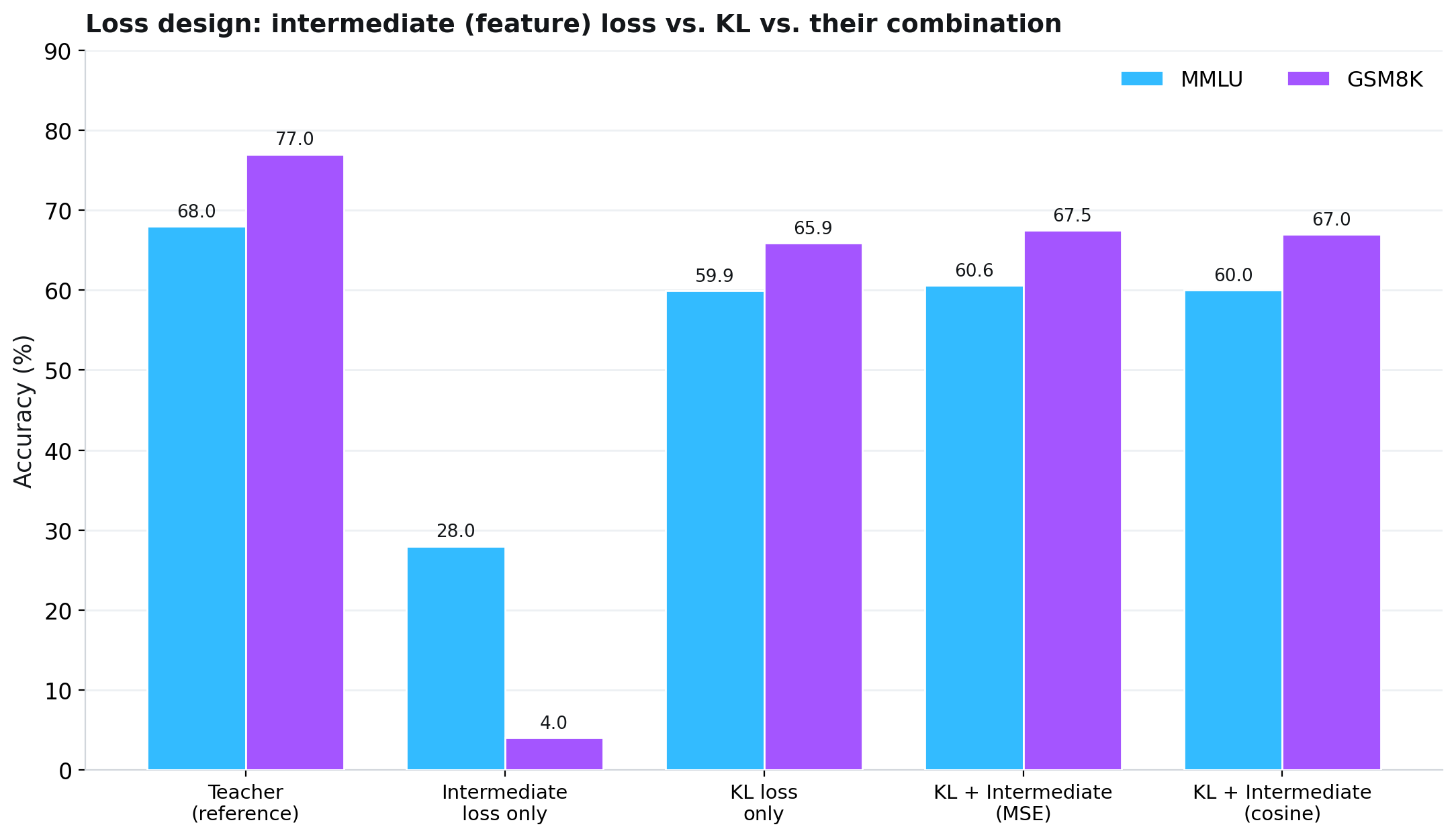}
  \caption{Loss design ablation. Best MMLU and GSM8K for a compact student
    healed under different losses, with the teacher for reference. An
    intermediate (feature) loss alone collapses; logit KL is indispensable; and
    logit KL plus a hidden-state feature loss is best.}
  \label{fig:loss-design}
\end{figure}

\textbf{Loss design.} Holding the student, teacher, and data budget fixed and
varying only the loss, the choice of loss is the dominant driver of recovery
(Figure~\ref{fig:loss-design}). An intermediate-layer feature loss
applied on its own collapses the student (MMLU \citep{hendrycks2021mmlu} near
$28\%$, GSM8K \citep{cobbe2021gsm8k} near $4\%$):
logit-level KL is indispensable, recovering MMLU to $59.9\%$ and GSM8K to
$65.9\%$. Adding a hidden-state feature loss on top of logit KL gives a small,
consistent gain, reaching $60.6\%$ MMLU and $67.5\%$ GSM8K (mean-squared-error
variant; a cosine variant is comparable). The recommendation is therefore
simple: always include logit KL, and add a hidden-state feature loss for a
reliable improvement.

\begin{figure}[t]
  \centering
  \includegraphics[width=0.85\linewidth]{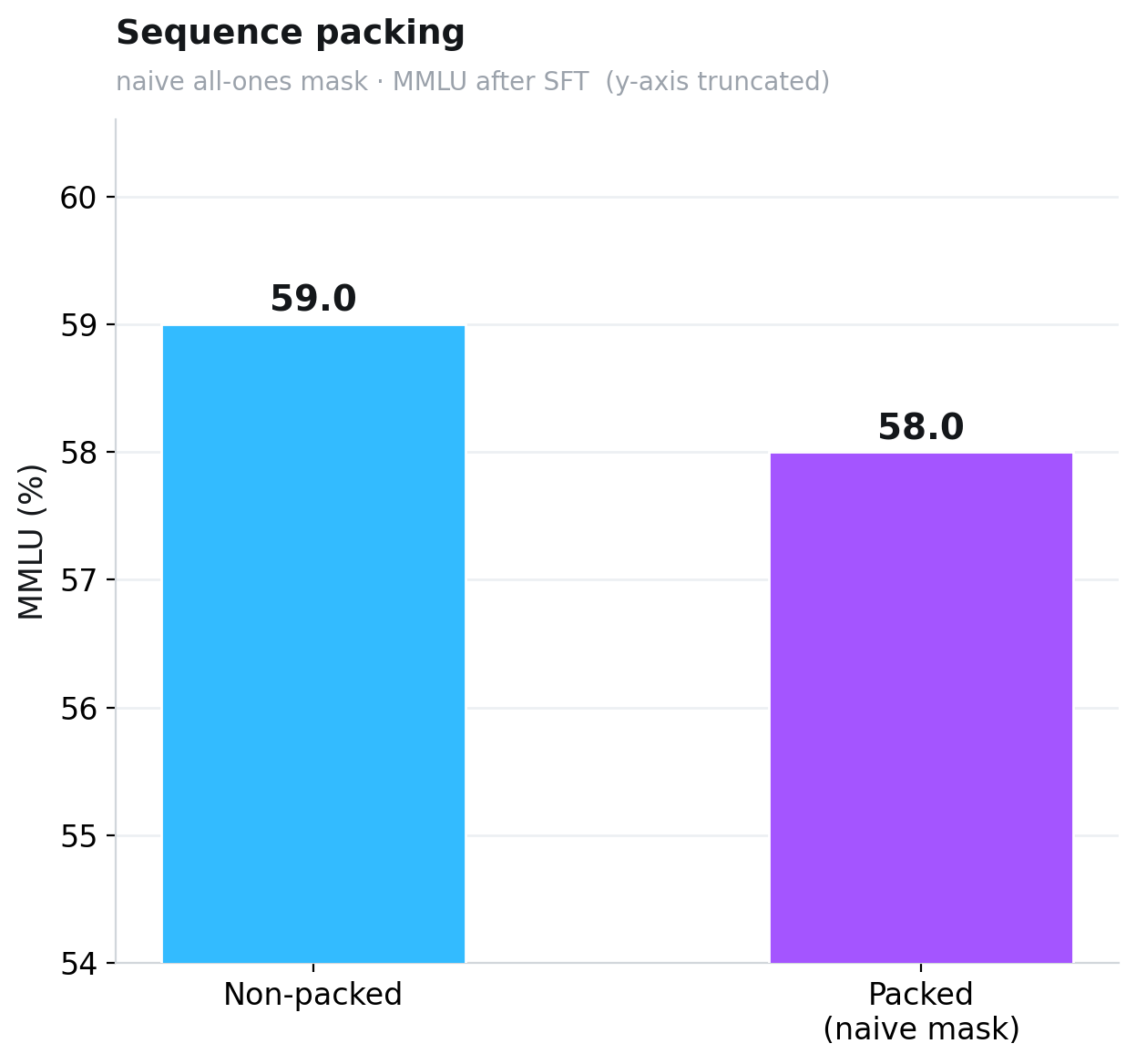}
  \caption{Sequence packing. MMLU after supervised fine-tuning for non-packed
    vs.\ naively packed (all-ones mask) training; packing costs about one
    point.}
  \label{fig:seq-packing}
\end{figure}

\textbf{Sequence packing.} Packing concatenates short examples into one long
sequence to keep the affordable long context full of useful tokens. Packing
with a naive all-ones attention mask, which permits attention across example
boundaries, costs only about one point of MMLU relative to non-packed training
(Figure~\ref{fig:seq-packing}). The teacher's KL signal appears to
compensate for the missing per-example block mask, so cheap naive packing is a
reasonable default for distillation.

\begin{figure}[t]
  \centering
  \includegraphics[width=\linewidth]{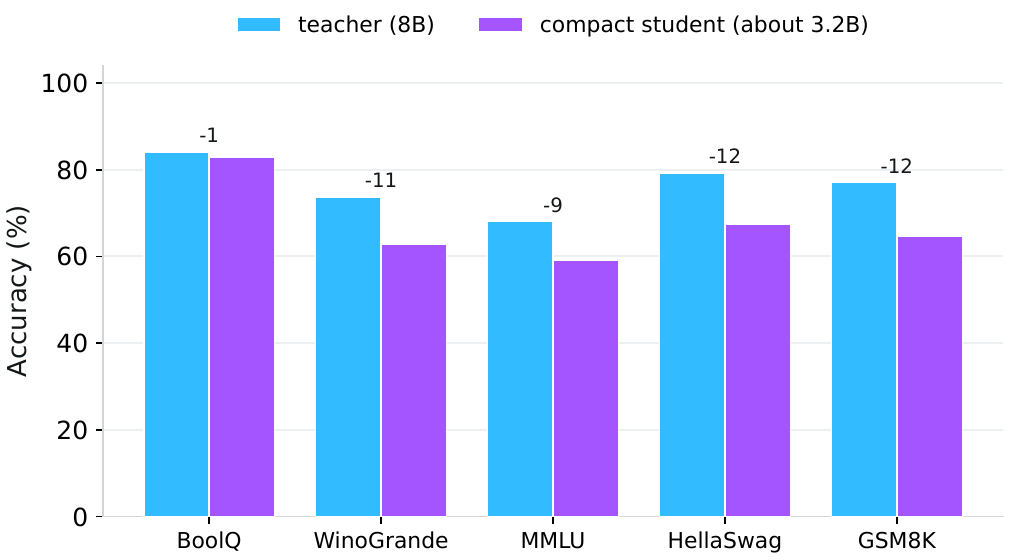}
  \caption{Short-context accuracy of the compact student against the teacher.
    The student retains most of the teacher's BoolQ and HellaSwag accuracy and
    stays within about nine points on MMLU, with larger gaps on WinoGrande and
    GSM8K.}
  \label{fig:compact-vs-teacher}
\end{figure}

Figure~\ref{fig:compact-vs-teacher} summarises the resulting compact student
against its teacher on BoolQ \citep{clark2019boolq}, WinoGrande
\citep{sakaguchi2020winogrande}, MMLU, HellaSwag \citep{zellers2019hellaswag},
and GSM8K: the student retains most of the teacher's short-context accuracy at
less than half the size.

\section{Limitations}
\label{sec:limitations}

Our study focuses on a single teacher--student pair: an 8B instruction-tuned
teacher and a compact $\sim$3.2B student. Although the recipe is intended to be
broadly applicable, we do not evaluate different model families, compression
methods, or student sizes, so the extent to which the recommendations transfer
to substantially different architectures remains open.

The 4K--256K loss-kernel sweep intentionally uses a toy output-projection
network with synthetic inputs. It isolates the asymptotic memory and timing of
the loss implementations, but it does not measure end-to-end training speed,
model quality, convergence, or interactions with attention and optimiser
state at those sequence lengths. We therefore use it as mechanistic evidence
alongside, not as a replacement for, the real-LLM experiments.


Our systems results are obtained with Megatron-Bridge and ModelOpt on
H200 GPUs. The fused chunked KL formulation is generic, but its efficiency
characteristics on other hardware and frameworks remain to be validated.


\section{Conclusion}
\label{sec:conclusion}
We presented a practitioner's recipe for efficient knowledge-distillation
recovery of a compact LLM. Two systems choices carry most of the benefit:
distil \emph{offline} from cached top-$K$ teacher logits to match online
quality at lower memory and higher throughput, and use a \emph{fused chunked KL
loss} so that peak memory is linear in sequence length and long-context healing
fits on a single GPU. A controlled loss-only benchmark isolates this mechanism:
the fully chunked implementation remains within $11.6$~GiB per GPU at 256K
tokens and overtakes the forward-chunked implementation in iteration rate at
long sequence lengths, while we explicitly separate these toy-network results
from end-to-end LLM throughput. Supporting ablations recommend combining logit
KL with a hidden-state feature loss and show that cheap naive sequence packing
costs only about a point of MMLU. The net result is most of the teacher's
short-context quality at a fraction of the size and training cost, with known
long-context limits. We made our chunked-loss implementation open-source. 

\bibliography{references}

\appendix
\newpage

\section{Experimental Configuration}
\label{app:config}

Table~\ref{tab:config} lists the configuration shared by all profiling runs
in \S\ref{sec:exp-online-offline} and \S\ref{sec:exp-chunked}, taken from the
merged configuration dumps recorded in the run logs. The runs vary only the
KL-loss implementation and the sequence length ($8{,}192$ or $32{,}768$); all
other settings are held fixed. Teacher top-$K$ logits for the offline runs
were precomputed with SGLang. Profiling used 15-iteration runs with NVIDIA
Nsight Systems capture over iterations 10--13 and per-step CUDA memory-history
snapshots.

\begin{table}[t]
\centering
\small
\begin{tabular}{@{}ll@{}}
\toprule
\multicolumn{2}{@{}l}{\emph{Student architecture ($3.23$B parameters)}}\\
\midrule
Layers / hidden / FFN size        & $28$ / $2{,}816$ / $7{,}168$ \\
Heads (GQA groups)                & $32$ ($8$), $128$ KV channels \\
Normalisation                     & RMSNorm ($\epsilon=10^{-5}$) \\
Positions / activation            & RoPE / SwiGLU, no biases \\
Dropout (attention / hidden)      & $0.0$ / $0.0$ \\
\midrule
\multicolumn{2}{@{}l}{\emph{Optimisation}}\\
\midrule
Optimiser                         & Adam ($0.9$, $0.999$, $10^{-8}$) \\
Weight decay / grad clip          & $0.1$ / $1.0$ \\
Learning rate (cosine)            & $2\!\times\!10^{-6}\!\rightarrow\!2\!\times\!10^{-7}$ \\
Warmup                            & $10$ iterations \\
Batch size (global / micro)       & $32$ / $1$ \\
Random seed                       & $1234$ \\
\midrule
\multicolumn{2}{@{}l}{\emph{Distillation}}\\
\midrule
Objective                         & forward KL, $\tau=1$ \\
Teacher support                   & top-$K$, $K=100$ \\
\midrule
\multicolumn{2}{@{}l}{\emph{Precision and parallelism}}\\
\midrule
Precision                         & bf16 mixed, FP32 grad reduction \\
Attention backend                 & FlashAttention \\
TP / PP / CP / EP                 & $1$ / $1$ / $1$ / $1$ (single GPU) \\
\midrule
\multicolumn{2}{@{}l}{\emph{Software stack}}\\
\midrule
Base container                    & NVIDIA NGC PyTorch 26.01 \\
                                  & (Ubuntu Linux, CUDA~12) \\
Training framework                & Megatron-Bridge v0.3.0 \\
                                  & (Megatron-Core, -FSDP) \\
                                  & ModelOpt \\
Kernels                           & Transformer Engine 2.11 \\
Multi-node comm.                  & NCCL 2.29, AWS EFA \\
                                  & (aws-ofi-nccl 1.18.0) \\
                                  & GDRCopy 2.5.1 \\
Other libraries                   & Transformers~4.57.6 \\
Teacher-logit precompute          & SGLang \\
\bottomrule
\end{tabular}
\caption{Training configuration shared by all profiling runs. TP/PP/CP/EP:
  tensor, pipeline, context, and expert parallelism.}
\label{tab:config}
\end{table}

\section{Toy Loss-Kernel Benchmark Configuration}
\label{app:toy-benchmark}
Table~\ref{tab:toy-config} records the controlled configuration used for
Figure~\ref{fig:loss-only-scaling}. The benchmark code constructs random but
deterministic hidden states, output-projection weights, and sparse teacher
targets once per configuration and presents equivalent inputs to all three
losses. It measures only the output projection and KL forward/backward path;
no transformer body or optimiser step is present. The plotted CSV selects a
$4{,}096$-token chunk from a sweep that also contains other chunk sizes.

\begin{table}[t]
\centering
\small
\begin{tabular}{@{}p{0.40\columnwidth}p{0.52\columnwidth}@{}}
\toprule
Component & Value \\
\midrule
Network scope & Vocabulary output projection only \\
Hidden / vocabulary size & $4{,}096$ / $131{,}072$ \\
Batch size / teacher support & $1$ / top-$100$ \\
Precision / temperature & bfloat16 / $1.0$ \\
Tensor parallelism & $2$ H200 GPUs \\
Plotted chunk size & $4{,}096$ tokens \\
Sequence lengths & 4K, 8K, 16K, 32K, 64K, 128K, 256K \\
Timing & Mean forward + backward after warm-up \\
Memory & Peak allocated CUDA memory per GPU \\
Isolation & Fresh distributed process per setting \\
\bottomrule
\end{tabular}
\caption{Configuration for the controlled loss-only benchmark. Timing and
  memory report the maximum rank value under tensor parallelism.}
\label{tab:toy-config}
\end{table}


\end{document}